\documentclass[10pt,twocolumn,letterpaper]{article}
\PassOptionsToPackage{table}{xcolor}
\usepackage[pagenumbers]{cvpr} 

\usepackage[dvipsnames]{xcolor}

\definecolor{mygrey}{rgb}{0.9,0.9,0.9}

\definecolor{cvprblue}{rgb}{0.21,0.49,0.74}
\usepackage[pagebackref,breaklinks,colorlinks,citecolor=cvprblue]{hyperref}
\usepackage{multirow}

\def\paperID{4829} 
\def\confName{CVPR}
\def\confYear{2024}

\title{Conditional Prompt Tuning for Multimodal Fusion}

\author{Ruixiang Jiang, \quad Lingbo Liu, \quad Changwen Chen \\
The Hong Kong Polytechnic University\\
Hong Kong SAR, China\\
{\tt\small rui-x.jiang@connect.polyu.hk}
\quad {\tt\small \{lingbo.liu, changwen.chen\}@polyu.edu.hk}
}

\begin{document}
\maketitle
\begin{abstract}

We show that the representation of one modality can effectively guide the prompting of another modality for parameter-efficient multimodal fusion. Specifically, we first encode one modality and use its representation as a prior to conditionally prompt all frozen layers of the other modality. This is achieved by disentangling the vanilla prompt vectors into three types of specialized prompts that adaptively capture global-level and instance-level features. To better produce the instance-wise prompt, we introduce the mixture of prompt experts (MoPE) to dynamically route each instance to the most suitable prompt experts for encoding. We further study a regularization term to avoid degenerated prompt expert routing. Thanks to our design, our method can effectively transfer the pretrained knowledge in unimodal encoders for downstream multimodal tasks. Compared with vanilla prompting, we show that our MoPE-based conditional prompting is more expressive, thereby scales better with training data and the total number of prompts. We also demonstrate that our prompt tuning is architecture-agnostic, thereby offering high modularity. Extensive experiments over three multimodal datasets demonstrate state-of-the-art results, matching or surpassing the performance achieved through fine-tuning, while only necessitating 0.7\% of the trainable parameters. Code will be released.

\end{abstract}    
\section{Introduction}
 
\begin{figure}
    \centering
    \includegraphics[width=0.75\linewidth]{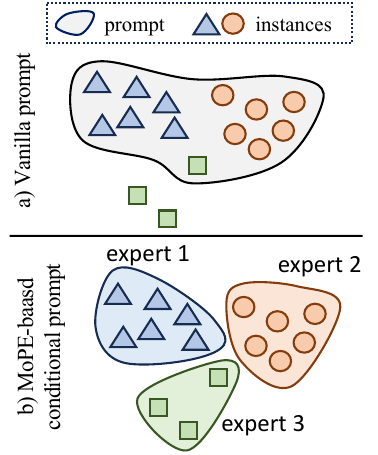}
    \caption{\textbf{High-level motivation of MoPE-based conditional prompting}. a) vanilla prompt-tuning learn a shared prompt for all instances, which would be influenced by majority classes; b) MoPE-based conditional prompt optimize multiple specialized prompt experts to better handle per-instance shift.}
    \label{fig:teaser}
\end{figure}

Empowered with billion-scale training data and highly scalable model architectures, recent unimodal pre-trained large models~\cite{radford2019language,liu2021swin,devlin2018bert,dosovitskiy2020image,kirillov2023segment}, also known as foundation models, have demonstrated their powerfulness that are transferrable to various downstream tasks~\cite{liu2021cross,ma2023segment,zhu2020incorporating}. Transferring multimodally-pretrained foundation models for a specific multimodal task, however, is less flexible. Recent explorations, such as CLIP~\cite{radford2021learning}, employ two tower designs to contrastively pre-train two encoders together. Despite its success, the joint pre-training entangles the two encoders, meaning that replacing either one would necessitate expensive paired pre-training from scratch. This limitation restricts the broader application of multimodal foundation models for downstream tasks that would benefit from a specific unimodal architecture. Therefore, a compelling question arises: \textit{Is it feasible to combine unimodal foundation models for downstream multimodal tasks?}

Two challenges emerge when transferring unimodal foundation models for multimodal task. The first challenge stems from the specialized architecture of foundation models across different modalities, each incorporating unique, modality-specific designs. This diversity complicates the design of an optimal fusion method for each combination of model architectures. Secondly, compared to unimodal datasets, the limited availability of multimodal data complicates the tuning of foundation models. As pre-trained models continue to increase in size, the standard fine-tuning technique may not be as efficient, given the large volumes of data needed to achieve optimal performance.

 We address above challenges by proposing a conditional prompt-tuning method. Specifically, we adopt a sequential pipeline to segregate the architecture details of each modalities. This design is architecture-agnostic, resulting in high modularity. To enhance the parameter-efficiency, we develop a fusion method based on the recently emerged prompt-tuning technique~\cite{li2021prefix,jia2022visual}, where we condition prompting of one modality on the other. Compared with fine-tuning, our method achieves similar or superior results while using significantly fewer trainable parameters.

Despite the effectiveness of prompt-tuning in various transfer learning setups, its application to multimodal tasks with large datasets is far from straightforward. Notably, prompt-tuning generally performs well in low-shot scenarios but can be less effective in full-shot training on the entire dataset~\cite{gao2022visual, liang2022modular,tsimpoukelli2021multimodal,yang2022prompt}. This reduced efficacy in large-scale scenarios could be explained by at least two factors: 1) Previous prompt-tuning methods optimize a globally-shared prompt for all instances~\cite{jia2022visual, li2021prefix}, which is not necessarily the local optimal for each input instance, and 2) the small amount of trainable parameters (compared with finetuning) can lead to underfitting on large datasets. Increasing the prompt length is a potential solution to incorporate more trainable parameters and better capture minority classes, but it often results in degraded outcomes instead~\cite{jia2022visual, yang2022prompt, khattak2023maple, li2021prefix,kim2023we}. In this paper, we aim to address these challenges by developing an instance-specific prompting technique and enhancing the expressiveness of vanilla prompt tuning.

To incorporate instance-specific prior into the framework of prompt-tuning, we augment the vanilla global-shared (\ie, static) prompt with instance-wise (\ie, dynamic) information to better handle per-instance shift. Specifically, we introduce two additional types of prompts: the \textit{dynamic prompt} augments the static prompt by capturing instance-level shift, while the \textit{mapped prompt} inject fine-grained information for multimodal prediction. We generate the two prompts in an instance-wise manner, utilizing two encoders for each modality in a sequential pipeline, where the output of the complementary modality is used to guide the prompting of the main modality encoder. Such design segregates the architectural details of encoders from each other, thereby permitting each modality to be independently substituted.

To further scale up the expressiveness of prompting, our key motivation is to fit multiple prompts to deal with the shift across instances, as illustrated in Fig.~\ref{fig:teaser}.  To be more specific, we introduce the \textbf{M}ixture \textbf{o}f \textbf{P}rompt \textbf{E}xpert (MoPE) method for dynamic prompt generation, where a pool of prompt experts is optimized per-layer. For a specific instance, we utilize a learned router to predict a routing score, which is used to weigh all experts for synthesizing the dynamic prompt for this specific instance. We also study the effect of a regularization term, which prevents degenerated routing across instances. With the combination of those modules, we observe specialized experts spontaneously emerge after training, each focusing on a specific group of instances. Moreover, we demonstrate that increasing the number of experts could be an effective way to scale up the model capacity compared with simply increasing prompt length. This allows for learning more complex mapping for down-stream tasks with a large dataset, while still keeping fixed sequence length during self-attention. In conclusion, we summarize our contribution as follows: 
\begin{itemize}
    \item We propose to augment the vanilla prompt tuning with the dynamic and mapped prompt, utilizing the paired modality as a prior, to better adapt the pretrained model to each instance.
    \item We elaborate a mixture of prompt expert design for dynamic prompt, which scales up the expressiveness of prompt tuning for transfer-learning.
    \item We study the effect of a regularization term for avoiding degenerated expert routing. 
    \item Extensive experiments on UPMC\_Food-101, SNLI-VE and MM-IMDB dataset demonstrate state-of-the-art performance for multimodal fusion.
\end{itemize}

\section{Related Works}

\begin{figure*}
    \centering
    \includegraphics[width=.9\linewidth]{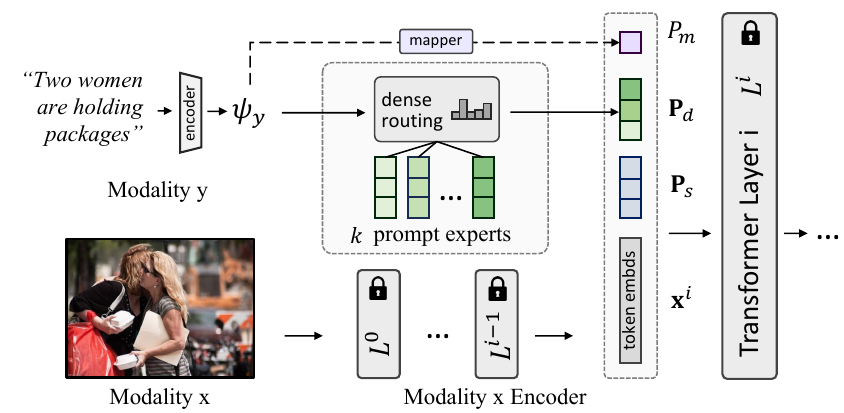}
    \caption{\textbf{Overview of instance-wise conditional prompting.} We illustrate the proposed conditional prompt tuning method for multimodal fusion, applied to one Transformer layer. Three types of prompts are concatenated to the token embeddings: (1) We first extract features from the complimentary modality $\psi_y$, which is used to guide the mixture of expert routing of instance-wise dynamic prompt $\mathbf{P}_d$. (2) A lightweight mapper is applied to map $\psi_y$ into a single prompt $P_m$. (3) the static prompt $\mathbf{P}_s$ will also be used, which is not conditional on $\psi_y$. All of the pretrained Transformer parameters are frozen, and only the router, mapper and prompt are trainable.}
    \label{fig:enter-label}
\end{figure*}

\textbf{Unimodal Prompt Tuning.} Prompt tuning emerges as a parameter-efficient method of transferring large pretrained models for down-stream tasks. This involves freezing the pretrained model and inserting additional learnable prompt tokens into the input of the model for solving specific tasks. This tuning scheme was first popularized in natural language processing (NLP)~\cite{li2021prefix, lester2021power}, then quickly introduced to computer vision (CV) society~\cite{jia2022visual, bahng2022exploring,liu2022prompt,lin2023being}. For either modality, this tuning scheme achieves good transfer learning performance in low-data regime, yet its performance is less comparable to fine-tuning when abundant training instances are available~\cite{li2021prefix,gao2022visual}. Moreover, simply increasing prompt length could result in performance saturation, and over-length prompts might have an negative impact on performance~\cite{yang2022prompt,jia2022visual,kim2023we}. Following works scale up the vanilla prompt tuning by increasing their diversity through chain-of-thought~\cite{zhang2022automatic}, adaptive prompting~\cite{asai2022attentional,wang2022learning,yang2023dynamic,huang2023diversity}, or ensembling~\cite{qin2021learning}. In this work, we scale up the expressiveness of vanilla prompting with a MoE design.

\textbf{Multimodal Prompt Tuning.} There are two prominent streams of work in multimodal prompting, depending on whether pre-training of the foundation model is multimodal or unimodal. The first stream applies prompt tuning to transfer \textit{multimodal pretrained models} for downstream tasks, primarily focusing on Vision-Language Models (VLMs) such as CLIP~\cite{radford2021learning}. Existing methods of this stream either optimize prompts on one branch (\eg, CoOp~\cite{zhou2022learning}, CPT~\cite{yao2021cpt}), or tune both branches with different designs~\cite{jiang2023clip,khattak2023maple,zhou2022conditional,zang2022unified}. More related to our work is the second stream, which focus on using prompt to bridge\textit{ unimodal pretrained models} for multimodal task. Frozen~\cite{tsimpoukelli2021multimodal} first introduces a method where the visual representation are mapped as two input tokens to query frozen language models (LM). PromptFuse and BlindPrompt~\cite{liang2022modular} improved upon this by introducing tunable prompts to the LM only for alignment. PICa~\cite{yang2022empirical} translate images into text captions to prompt frozen Large LMs. PMF~\cite{li2023efficient} further introduces interactive prompt-tuning only in the deep layers for memory efficiency. Our work extends the second branch by further studying the interplay of prompting across modalities.

\textbf{MoE in Transformers.} MoE is a powerful technique for scaling  Transformers up to billion scale~\cite{lepikhin2020gshard,riquelme2021scaling,mustafa2022multimodal}. The fundamental approach involves inserting MoE layers into the standard Transformer architecture, composed of multiple feed-forward networks (FFNs) acting as experts. A router is learned to route each token embedding to the most suitable expert(s). Due to the significant computational cost associated with FFN forwarding, a sparse gating function is typically employed to limit the number of experts used per token~\cite{qin2021learning,riquelme2021scaling}.  Thanks to those designs, the MoE layer can efficiently scale up model capacity. Inspired by previous FFN-based MoE, we explore scaling up the prompt tuning method with an MoE design.

\section{Method}
In this section, we elaborate on the proposed method for multimodal fusion via conditional prompting. In Sec.~\ref{sec: prelim} we establish the basic notation of prompt tuning. We explain how we sequentially fuse two unimodal encoders via prompting, as well as the three types of prompts in Sec.~\ref{sec: cpt}. In Sec.~\ref{sec: mop} we introduce the MoPE-based conditional prompting technique. Finally, we introduce the regularization method in Sec.~\ref{sec: reg}.

\begin{figure*}
    \centering
    \includegraphics[width=0.9\linewidth]{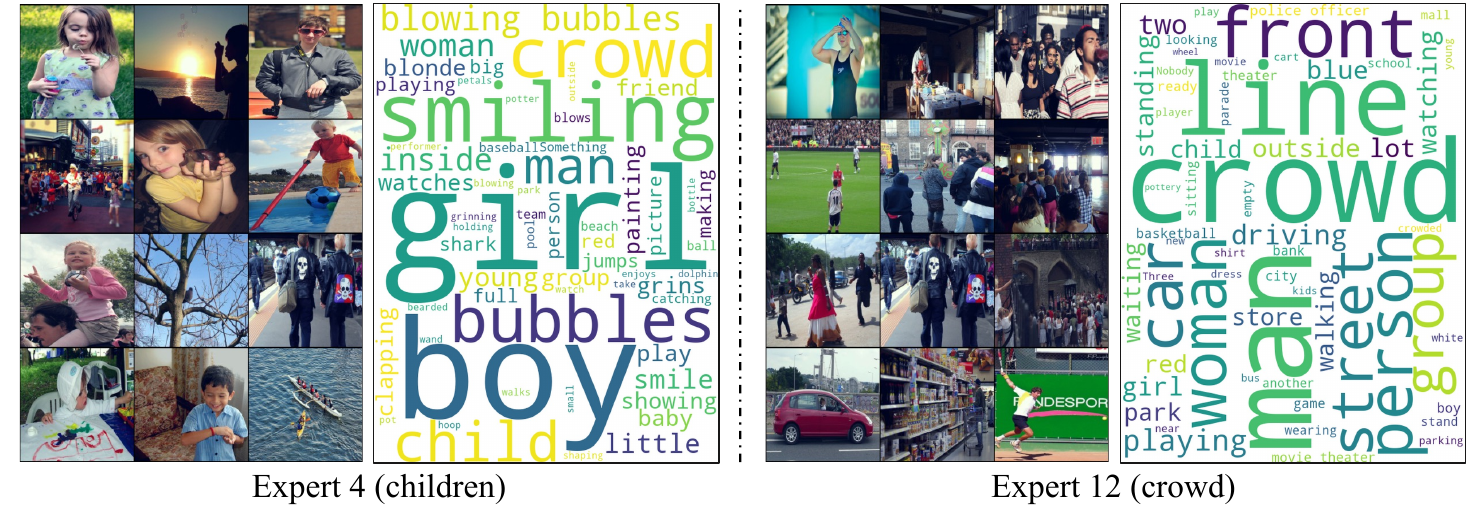}
    \caption{\textbf{Examples of expert routing.} We visualize the result of last-layer expert routing on the SNLI-VE test set, by showing the images (left) and the paired word clouds (right) routed to expert 4 and 12. The expert ID of an instance is determined by its highest routing score for visualization purposes only. In this example, expert-4 is specialized for children while expert-12 focuses on crowds.}
    \label{fig:expert}
\end{figure*}

\subsection{Preliminary: Prompt Tuning}\label{sec: prelim}

Consider using a Transformer~\cite{vaswani2017attention} or its variants to extract features from (embedded) input sequence $\mathbf{x}^0 \in \mathbb{R}_{s\times d}$ where $s$ is the sequence length and $d$ is hidden dimension of the Transformer. The input sequence of $i$-th layer $L^i$ could be further denoted as:

\begin{equation}
    \mathbf{x}^i = [x^i_0,\mathbf{T}^{i-1}]
\end{equation}
where $x^i_0$ denote the [CLS] token, $\mathbf{T}^{i-1}$ is the token embedding from the previous layer, and $[, ]$ denotes the concatenation operation.  


Instead of fine-tuning all of the parameters, prompt-tuning freezes all pre-trained model weights and insert a small amount of trainable prompts $\mathbf{P}\in \mathbb{R}_{l\times d}$ to the input $\mathbf{x}^0$ or internal layers $\mathbf{x}^i$, where $l$ is the number of prompts. The input of layer $L^i$ now becomes:

\begin{equation}
    \mathbf{\hat{x}}^i = [x^i_0,\mathbf{P},\mathbf{T}^{i-1}]
\end{equation}

In the self-attention process, the learned prompt $\mathbf{P}$ attends to the whole sequence $\mathbf{\hat{x}}^i$ to achieve parameter-efficient transfer learning.  Unless otherwise specified, we assume that prompt-tuning append learnable prompts to all of the transformer layers, this corresponds to ``VPT-deep'' in CV and ``prefix-tuning'' in NLP.

\subsection{Conditional Prompt Tuning} \label{sec: cpt}


Our objective is to transfer unimodal pretrained models for solving multimodal tasks via prompt tuning. With paired multimodal input data as a prior, we aim to condition the prompting of one modality on the other for more effective and adaptive transfer learning. To achieve this goal, we disentangle the vanilla prompt vector $\mathbf{P}$ into three types of specialized prompts $[\mathbf{P}_s, \mathbf{P}_d, P_m]$. The static prompt $\mathbf{P}_s \in \mathbb{R}_{l\times d}$ is a globally-shared prompt vector that is non-conditional to input. The dynamic prompt $\mathbf{P}_d\in \mathbb{R}_{l\times d}$ and the mapped prompt $P_m\in \mathbb{R}_{d}$ are instance-wise adaptive, which are conditionally synthesized based on the representation of the other modality(ies). 

A sequential pipeline is adopted to achieve prompt conditioning. More specifically, let $(x, y)$ be a multimodal input pair, with $\mathcal{E}_x$, $\mathcal{E}_y$ be the encoders of each modality. Depending on the intrinsic of the specific task, we assign a fusion direction from the complementary modality to the main modality, and the proposed prompt tuning method will only be used for tuning the main modality. Without loss of generality, we treat $x$ as the main modality, so the $y$ would be complementary. To synthesize the instance-wise prompt vector $\mathbf{P}_d$ and $P_m$, we first extract the global-level feature of complementary modality $\psi_y = \mathcal{E}_y(y)$ and apply a router $R(\cdot)$ to synthesize the instance-wise dynamic prompt vectors $\mathbf{P}_d$ to tune $\mathcal{E}_x$ (entailed in Sec~\ref{sec: mop}). To further inject information from the complimentary modality, we apply a lightweight mapper to map the complimentary feature as a single prompt: $P_m = f_m(\psi_y)$. Finally, the static prompt $\mathbf{P}_s$ will also be used to tune the main modality encoder $\mathcal{E}_x$. In summary, the input of layer $L^i$ of $\mathcal{E}_x$ becomes:

\begin{equation}
    \mathbf{\hat{x}}^i = [x^i_0,\mathbf{P}_s,R(\psi_y), f_m(\psi_y),\mathbf{T}^{i-1}]
\end{equation}

It is important to note that our three types of prompting serve as a plug-in module to replace the vanilla prompt vector of the main modality $\mathcal{E}_x$. We do not impose any constraints on the architecture, capacity, or tuning method of the complementary modality $\mathcal{E}_y$.

\subsection{Mixture of Prompt Experts}\label{sec: mop}
We propose learning multiple prompts as experts for generating the dynamic prompt in a per-instance manner to handle per-instance shift. Specifically, For each prompt-tuned Transformer layer $L^i$, we randomly initialize $k$ prompt experts $\{\mathbf{E}_i\}_{i=1}^k$ to be optimized end-to-end, where $\mathbf{E}_i\in\mathbb{R}_{l\times d}$. For a specific instance, its dynamic prompt at this layer would be synthesized based on all of the available experts. This is accomplished by learning a $\operatorname{Softmax}$ layer to predict a routing score $r = \operatorname{Softmax}(\mathbf{W}_r\psi_y / \tau + \epsilon)$, where $\mathbf{W}_r$ is layer-specific linear transformation, $\tau = 0.1$ is the temperature hyper-parameter, and $\epsilon$ is sampled noise to encourage routing diversity~\cite{riquelme2021scaling}. The routing process $R(\psi_y)$ is to synthesize the dynamic prompt by a convex combination of all experts according to the routing score:
\begin{equation}
\mathbf{P}_d = \sum_i^k r_i \mathbf{E}_i
\end{equation}

Unlike previous MLP-based MoE methods~\cite{riquelme2021scaling}, we do not insist the routing score be sparse (\ie, we do not use a TOP-K gate). This is because all prompts are the leaf nodes in the computation graph, and densely fusing them does not result in high computational cost, while we empirically find it gives better results.






\subsection{Regularizing Expert Routing} \label{sec: reg}
 To avoid specific experts being dominant across all instances, we add an additional importance loss~\cite{riquelme2021scaling} to encourage balanced expert routing. Specifically, for a batch of input $\mathbf{Y}$, the importance of expert-\textit{i} is defined as the summed routing score of all inputs.

\begin{equation}
    \operatorname{Imp}(E_i) = \sum_{y\in \mathbf{Y}} \operatorname{Softmax}(\mathbf{W}_r\psi_y / \tau)_i
\end{equation}
The importance loss is defined as coefficient of variation of all experts' importance:

\begin{equation}
    \mathcal{L}_{imp} = \operatorname{stopgrad}\left( \left(\frac{\operatorname{std}(\{\operatorname{Imp}(\mathbf{E}_i)\}_i^k)}{\operatorname{mean}(\{\operatorname{Imp}(\mathbf{E}_i)\}_i^k)} \right)^2; \gamma \right)
\end{equation}

where $\operatorname{stopgrad}(\cdot)$ is the stop-gradient operator, which prevents error propagation of this loss term when the coefficient of variation is less than a pre-defined threshold $\gamma = 0.05$. This loss encourages balanced utilization of experts across all instances. The inclusion of the additional threshold constraint is due to our instance-wise routing. Unlike the previous per-token routing, instance-wise routing uses a smaller batch size for importance value calculation, thereby increasing the likelihood of a larger coefficient of variation. 







\section{Experiment and Results}
\begin{table*}[]

\centering
\resizebox{1.4\columnwidth}{!}{%
\begin{tabular}{l|lcccc}
\toprule
&Method       & Param & SNLI-VE $\uparrow$ & Food-101 $\uparrow$ & MM-IMDB $\uparrow$     \\
\hline \hline                                                  
\multirow{5}{*}{\rotatebox[origin=c]{90}{\textit{finetuning}}} & ImgOnly      &  86.9M     &  33.37       &    75.84       &   39.31/53.85           \\
&TextOnly     &   109.0M    &  69.69      &  86.46  &      58.80/65.67 \\
&LateConcat   &   196.0M    & 70.01   & 93.29    & 59.56/64.92  \\
&SequentialFuse &  197.0M   & 74.44      &  87.38        &59.53/66.55\\                    
&MMBT~\cite{kiela2019supervised}         &   196.5M    & 67.58   & 94.10    & 60.80/66.10 \\\cline{1-6}
\multirow{11}{*}{\rotatebox[origin=c]{90}{\textit{prompt-tuning}}}  & P-ImgOnly      &   0.1M    &  33.30       &    75.84       &   32.83/49.47           \\
&P-TextOnly     &   -    &  69.69      &  81.20  &      51.84/61.81 \\
&P-LateConcat &    1.0M   & 63.05   & 89.03    & 53.91/59.93 \\
&P-SequentialFuse & 1.1M  & 65.26 & 81.50 & 55.57/63.98 \\
&P-MMBT~\cite{li2023efficient}       &   0.9M    & 67.58   & 81.07    & 52.95/59.30 \\
&PromptFuse~\cite{liang2022modular}   & -      & 64.53   & 82.21    & 48.59/54.49 \\
&BlindPrompt~\cite{liang2022modular}  & -      & 65.54   & 84.56    & 50.18/56.46 \\
&PMF~\cite{li2023efficient}          & 2.5M  & 71.92   & 91.51    & 58.77/64.51 \\ 
&PMF-Large~\cite{li2023efficient}          & 4.5M  & 72.10   & 91.68    & 61.66/66.58 \\ 

&\cellcolor{mygrey}Ours ($k=4$)         & \cellcolor{mygrey}1.5M & \cellcolor{mygrey}73.32   & \cellcolor{mygrey}91.54       & \cellcolor{mygrey}61.54/67.49\\
&\cellcolor{mygrey}Ours ($k=16$)         & \cellcolor{mygrey}2.6M & \cellcolor{mygrey}\textbf{73.85}   &\cellcolor{mygrey}\textbf{91.74}       & \cellcolor{mygrey}\textbf{62.01}/\textbf{68.25}\\
\bottomrule
\end{tabular}

}
\caption{\textbf{Quantitative results on three multimodal dataset.} Quantitative result of all of the baseline methods, compared methods, and our method with different expert numbers. The metric on SNLI-VE and Food-101 is accuracy (\%), and MM-IMDB is F1-Macro and F1-Micro. We also list the total number of trainable parameters (million) of each method, where `-` means parameter less than 0.1 million. The best prompt-based results are in bold.}
\label{tab: classfication}
\end{table*}

\subsection{Implementation Details}
\textbf{Architecture Details.} For all experiments, unless otherwise specified, we use Swin-B~\cite{liu2021swin} as the vision encoder and Bert-base~\cite{devlin2018bert} as the text encoder. Following the experiment setup in~\cite{jia2022visual,li2023efficient}, we also finetune a linear head for each dataset. We implement the mapper $f_m(\cdot)$ as a two-layer MLP with GELU nonlinearily. Regarding prompt tuning, we use $l=6$ prompts and $k=16$ experts by default, and the tunable prompt is applied to all layers of the main modality encoder. We employ vanilla prompt tuning~\cite{jia2022visual} to tune the complementary modality.

\textbf{Training Detail.} All images are resized and cropped to the size of $224\times 224$. All models are trained for 12 epoches, using the AdamW~\cite{loshchilov2017decoupled} optimizer with a learning rate of $4e-4$ for vision and $5e-4$ for text. We use RandAug~\cite{cubuk2020randaugment} with default setup to augment the training images, with the exception of the SNLI-VE dataset.  We use a constant step decay scheduler that halves the learning rate at epoch = 2 and 5.

\subsection{Datasets}

\textbf{UPMC\_Food101}~\cite{wang2015recipe} serves as a comprehensive multimodal dataset designed for fine-grained recipe classification.  The dataset contains 90,840 image-text pairs for 101 food classes. We follow previous methods~\cite{li2023efficient,kiela2019supervised} to create a validation set of 5000 samples.

\textbf{MM-IMDB}~\cite{arevalo2017gated} is a multimodal movie classification dataset.  It comprises 25,956 pairs of images and texts, each pair including a movie poster and a plot summary. The dataset supports multi-label genre classification across a spectrum of 23 genres with imbalanced classes.

\textbf{SNLI-VE}~\cite{xie2019visual} is a large-scale multimodal dataset with 565,286 image-text pairs. The task for this dataset is visual entailment, which means that the model should decide whether a hypothesis matches the given premise. Following~\cite{li2023efficient}, we only use the image premise, while in other works the text premise might also be used.

\subsection{Compared Methods}
First, we compare our method with several baseline methods for multimodal fusion.

    \textbf{ImgOnly / TextOnly}. Only fine-tune one encoder, and the input of the other modality is discarded.

    \textbf{P-ImgOnly / P-TextOnly}. Only prompt-tune one encoder, and the input of the other modality is discarded. In our experiment, we use VPT-deep~\cite{jia2022visual} style prompt-tuning for the vision and text transformer.

    \textbf{LateConcat}.  This baseline involves fine-tuning both encoders, concatenating their features, and appending a linear head for classification.

    \textbf{P-LateConcat}. Similar as \textbf{LateConcat} but prompt-tune each encoder instead of fine-tuning. However, the linear head is still fine-tuned.

    \textbf{SequentialFuse}. This method first extracts features from the complementary modality and maps them to the embedding space of the main modality encoder in the same way as our mapped prompt for end-to-end training. Both encoders are fine-tuned. This is a strong baseline and can be seen as our method without MoPE and routers, but with all parameters fine-tuned.

    \textbf{P-SequentialFuse}. Similar as \textbf{SequentialFuse} but prompt-tune each encoder. This baseline is comparable to CoCoOp~\cite{zhou2022conditional} but with additional static prompts.

To ensure a fair comparison, all the above methods with prompt-tuning use the same prompt length as our method. In addition to these baselines, we also compare our methods with existing prompt-based fusion methods, including MMBT~\cite{kiela2019supervised}, Frozen~\cite{tsimpoukelli2021multimodal}, PromptFuse~\cite{liang2022modular}, BlindPrompt~\cite{liang2022modular}, and PMF~\cite{li2023efficient}.

\begin{table}[]
    \centering
    \begin{tabular}{c|ccc}
    \toprule
         Prompt & SNLI-VE $\uparrow$  &Food-101 $\uparrow$& MM-IMDB $\uparrow$ \\ \hline \hline
         $[\mathbf{P}_s]$ &  33.30  & 75.2 &24.69/45.11 \\

        $[\mathbf{P}_d]$ &   64.52 & 74.22 & 46.90/60.17\\
        $[P_m]$ &  33.94  & 72.70 &24.69/45.11 \\
        $[\mathbf{P}_s, \mathbf{P}_d]$ & 66.92 &  74.24 & 48.26/60.96\\
        $[\mathbf{P}_s, P_m]$ & 65.26  & 81.40& 55.57/63.98 \\
        $[\mathbf{P}_d, P_m]$ & 71.81   &  91.13 & 60.42/66.88\\
        $[\mathbf{P}_s, \mathbf{P}_d, P_m]$&\textbf{73.85}  & \textbf{91.74} & \textbf{62.01}/\textbf{68.25}\\
        \bottomrule
    \end{tabular}
    \caption{\textbf{Ablation on each of the prompt types.} The results from all combinations of mapped prompt, dynamic prompt, and static prompt are presented. Our full method with all prompts achieves the best result.}
    \label{tab:ablate_prompt}
\end{table}

\subsection{Main Results}

The quantitative results of all baselines, compared methods, and our method with different expert number $k$ are summarized in Tab~\ref{tab: classfication}. We also report the total trainable parameters of all methods. 

Our method outperforms all the listed prompt-based methods and is competitive with fine-tuning. Specifically, when compared with the fine-tuning baselines, SequentialFuse and LateConcat, our method delivers competitive accuracy on the Food-101 dataset and superior results on the SNLI-VE and MM-IMDB datasets, while requiring as few as 0.7\% of the trainable parameters. Our method also matches the performance achieved by MMBT~\cite{kiela2019supervised}, an early fusion method, which suggests that the proposed sequential pipeline could be a promising fusion paradigm.

Compared to the prompt-tuning baseline, our method demonstrates superior parameter-efficiency. Our MoPE design enables us to achieve significant gains of 12.34\%, 12.32\% and 10.75\%/5.48\% respectively on all metrics, outperforming P-SequentialFuse. We also surpass all existing prompt-based fusion methods, including PromptFuse~\cite{liang2022modular}, BlindPrompt~\cite{liang2022modular}, and PMF~\cite{li2023efficient}. Notably, our method with $k=4$ delivers similar results to PMF-Large, while only requiring 33\% of the trainable parameters, and our method with $k=16$ establishes a new state-of-the-art on the three datasets.

\subsection{Qualitative Result of MoPE Routing} Specialized prompt experts naturally emerge during training. We visualize two of them in Fig~\ref{fig:expert}. Specifically, we manually designate the expert with the highest routing score as the expert for each instance for visualization. By doing so, we are able to collect all text that would be routed to this expert, as well as the paired images. For each expert, we create a word cloud of all the texts, with common stop words (\eg,\textit{``an'' ``the''}) removed. We also visualize some images that are paired with the text. In the provided example, we observe that expert-4 appears to specialize in children, while expert-12 focuses on crowds.

\section{Analysis and Discussion}

\textbf{Ablation: Effect of each prompt types.} We ablate each of the prompt types and report the metrics on three datasets, the result is summarized in Tab.~\ref{tab:ablate_prompt}. Our full method with all types of prompts achieves the best result, indicating that each prompt collaborates with the others. Specifically, without the mapped prompt, the fine-grained information from complimentary modality is lost, leading to a significant drop of 10.3\%, 23.6\% and 28.4\%/11.9\% on all metrics, respectively.
When the dynamic prompt is not used, the method simply find a global prompt, which may not handle every instance well. This is supported by a drop of 11.6\% F1-Macro, which is known to be sensitive to class-imbalance.
Without the static prompt, the model already achieves satisfactory result, due to the fact that dynamic prompts with a degenerated router could also act as static to learn feature across all instances. However, adding the static prompt relieves the stress of dynamic prompts for global feature modeling, allowing them to focus more on capturing instance-level features. This leads to an improvement of 2.84\%, 0.67\%, and 2.63\%/2.04\%, respectively.

\begin{figure}
    \centering
    \includegraphics[width=\linewidth]{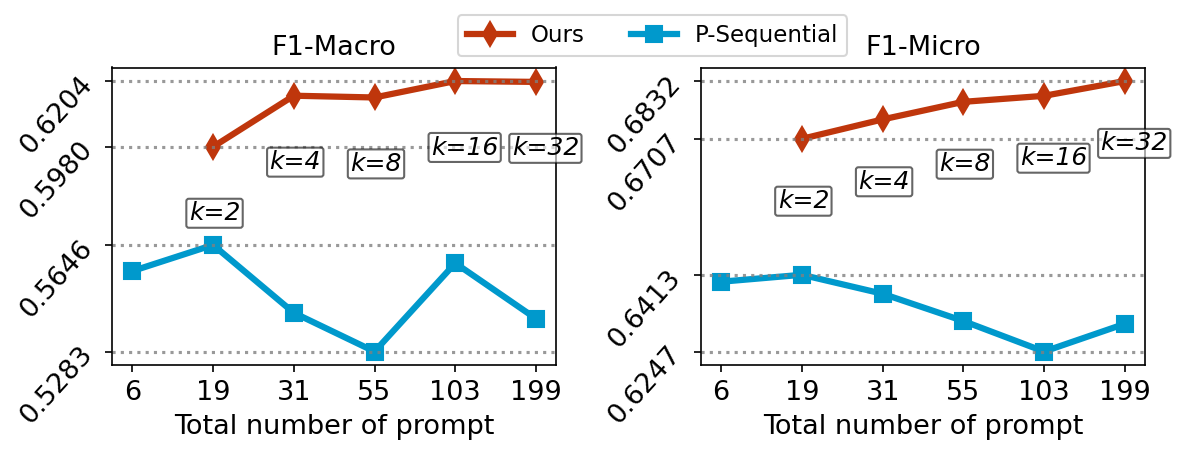}
    \caption{\textbf{More experts v.s. longer prompts.} We compare the effects of increasing the number of experts, $k$, versus increasing the prompt length, $l$. Increasing the number of experts consistently outperforms the strategy of lengthening the prompts, exhibiting a linear growth trend.  Conversely, excessive prompt length detrimentally affect the model's performance}
    \label{fig:ablate_k}
\end{figure}

\textbf{Ablation: Effect of MoPE.} We study the effectiveness of MoPE by increasing the number of experts $k$ v.s. scaling up prompt length $l$. Specifically, our starting point is $l=6$ prompts and $k = 2$ experts, which account for $(2+1)\times6+1 = 19$ tunable prompts in total per-layer. We increase the number of $k$, and at each step we also report the result of simply increasing $l$ to the same total number of prompts. The results, measured as F1-Macro and F1-Micro on the MM-IMDB dataset, are summarised in Fig.~\ref{fig:ablate_k}.

For both metrics, our results indicate that merely augmenting the number of prompts $l$ does not lead to a steady performance improvement. Such observation is in-line with previous studies~~\cite{jia2022visual, yang2022prompt, khattak2023maple, li2021prefix}. Moreover, the increase in prompt length will slow down the model due to the quadratic time complexity of self-attention. On the other hand, our MoPE scales better with the total number of prompts, as indicated by the linear growth of performance with respect to total trainable prompts. This improvement is achieved by conditioning the dynamic prompt on more experts to enhance the prompt's expressiveness. Since the number of prompts used in the forward pass remains fixed, our method maintains a constant time complexity.

\begin{figure}
    \centering
    \includegraphics[width=\linewidth]{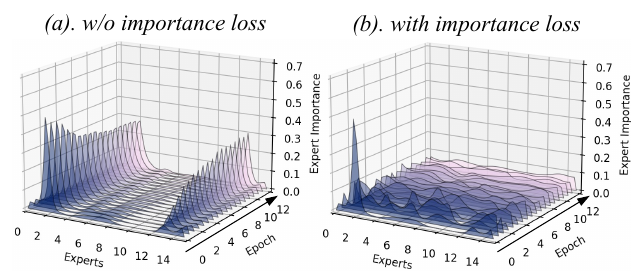}
    \caption{\textbf{Effect of the importance loss}. We visualize how importance (z-axis) of all experts (x-axis) in the last Transformer layer change during training (y-axis). (a) without importance loss, only a few experts are used throughout training (b) the importance loss ensures balanced utilization of all experts.}
    \label{fig:route-time}
\end{figure}

\textbf{Ablation: Effect of the importance loss.}
The importance loss is crucial for avoiding degenerated routing solution. In Fig.~\ref{fig:route-time}, we visualize how the importance of each expert (\ie, average routing score) changes when training on the SNLI-VE dataset. Without the importance loss, routing adheres to its initial state, leading to a skewed distribution where few experts dominate. This phenomenon aligns with observations from previous FFN-based MoE methods~\cite{eigen2013learning,shazeer2017outrageously}. The importance loss alleviates this by penalizing unbalanced expert importance, which results in balanced expert utilization of all experts.

\textbf{Ablation: Dense routing vs. sparse routing.}
Previous MoE with FFN experts usually employ a sparse gating function that only select one or a few experts for token, while we use a dense routing to allow all experts to make contributions. To compare the effect, we also implement a sparse router following~\cite{riquelme2021scaling}, which takes form:
\begin{equation}
    r^\prime = \operatorname{TOP}_1( \operatorname{Softmax}(\mathbf{W}_r\psi_y / \tau + \epsilon))
\end{equation}

The results of using sparse routing versus dense routing are reported in Tab.~\ref{tab:sparse}. Our experiments indicate that switching between either routing scheme does not induce significant performance gaps within the MoPE setting, while we favor the dense one as it gives marginally better results. 

\begin{figure}
    \centering
    \includegraphics[width=\linewidth]{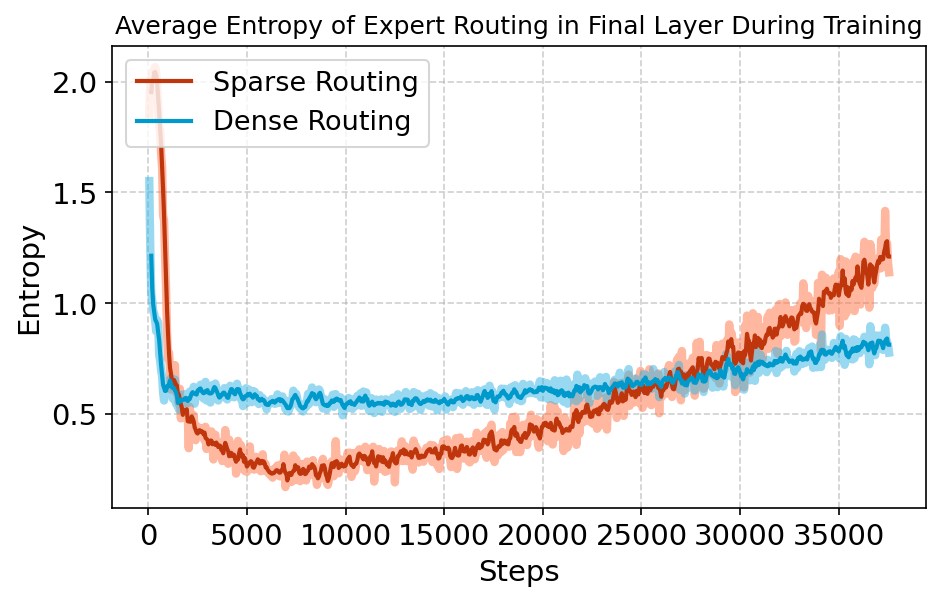}
    \caption{\textbf{Routing randomness of dense vs. sparse routing.} The average routing entropy throughout the optimization process for both dense and sparse routing is visualized. Dense routing exhibits less randomness}
    \label{fig:entropy}
\end{figure}

To further understand how dense routing helps the model, we evaluated the degree of uncertainty in routing, utilizing the empirical entropy of the routing score, averaged across training batches and all layers. Basically, high entropy indicate that routing behaves more randomly. The entropy is defined as $\mathcal{H}(r) = -\sum_{i=1}^nr_i log_2(r_i)$, where $r$ corresponds to the raw SoftMax probability regardless of the application of the TOP-1 gate. The relationship between entropy and optimization steps is illustrated in Fig.~\ref{fig:entropy}. As demonstrated in the figure, dense routing results in less random expert decision, suggesting that the model better leverages the complementary modality for routing.

\begin{table}[]
    \centering
    \begin{tabular}{c|cccc}
    \toprule
         Routing& SNLI-VE $\uparrow$&Food-101 $\uparrow$ &MM-IMDB $\uparrow$ \\\hline \hline
         Dense & \textbf{73.85} & \textbf{91.74} & \textbf{62.01}/\textbf{68.25}\\
         Sparse & 73.00 &  91.20& 61.95/68.11\\
    \bottomrule
    \end{tabular}
    \caption{\textbf{Result of dense routing v.s. sparse routing}. Dense routing achieves slightly better results on all three datasets.}
    \label{tab:sparse}
\end{table}

\begin{figure}
    \centering
    \includegraphics[width=\linewidth]{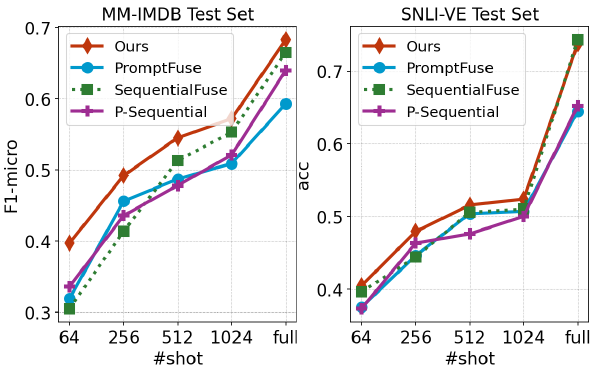}
    \caption{\textbf{Scaling performance with increased training data}. This chart showcases the comparative performance of our method and representative methods as we progressively increase the amount of training data, or 'shots'. The proposed method consistently achieves superior results, outperforming other prompt-tuning methods at all data scales, and remains competitive with the full-tuning method, \textit{PromptFuse}.  Note: \textit{P-Sequential} means \textit{P-SequentialFuse}.}
    \label{fig:n_shot}
\end{figure}

\textbf{Scalability with dataset size.} The performance of previous prompting method does not scale well with respect to increased training data~\cite{gao2022visual, liang2022modular,tsimpoukelli2021multimodal,yang2022prompt}. We study how the proposed method performs on different data scales. Specifically, we sub-sample the training set to 64, 256, 512, and 1024 shots, and train our method and other prompt-tuning and fine-tuning methods with the same subsampled data. The mean accuracy and F1-Micro score on SNLI-VE and MM-IMDB, respectively, are reported in Fig.~\ref{fig:n_shot}.

We observe that the two compared prompting methods, PromptFuse and P-SequentialFuse, provide satisfactory results in low-data situations. However, when a larger quantity of training samples is available, as is the case with 1024-shot and full-dataset training, they exhibit a significant performance gap compared to the fine-tuning method, SequentialFuse. In contrast, our method consistently matches or exceeds the results of fine-tuning across all data scales. We also note that our method significantly outperforms all compared methods on the low-shot MM-IMDB F1-micro metric. This superior performance is linked to the observation that the routing entropy is high in the low-shot regime. This suggests that our MoPE could adaptively behave like ensembling to prevent overfitting.

\textbf{Modularity.}  The proposed fusion method abstracts the complimentary modality as a representation, allowing high modality of both modalities. In particular, our method allow arbitrary models to be seamless plug-in for multimodal fusion. Such modularity is at least three fold: model architecture, the pre-training scheme, and the specific transfer learning method of the model. We exemplify each in Tab.~\ref{tab: modular}.

\textbf{Limitation and future work.} The proposed method relies on a single global-level representation for the complimentary modality, which might overlook the spatial information useful for specific tasks such as segmentation. Future works could study how to allow arbitrary sequence length from the complementary modality for fusion.

\begin{table}[]
    \centering
    \begin{tabular}{ccccc}
    \toprule
         Architecture& Pretraining & Transfer &MM-IMDB $\uparrow$ \\ \hline \hline
         BoW& Bert\textsuperscript{*}& Fine-tuning& 48.20/57.50\\
       Transformer &Bert~\cite{devlin2018bert}& Frozen & 58.86/66.13\\
         Transformer &GPT-2~\cite{radford2019language}& Frozen& 34.03/50.84\\
         Transformer &Bert~\cite{devlin2018bert}& Fine-tuning& 60.34/67.27\\

    \bottomrule
    \end{tabular}
    \caption{\textbf{Our prompt tuning method are highly modular.} We offer flexibility in at least three dimensions: model architecture, pretraining scheme, as well as transfer learning technique. (*): Bag-of-word initialized with Bert word embeddings.}
    \label{tab: modular}
\end{table}


\section{Conclusion}
In this paper, we present a conditional prompting method for parameter-efficient multimodal fusion. Our method involves augmenting the vanilla static prompt with dynamic and mapped prompt for instance-adaptive prompt learning. We also introduce the mixture-of-prompt-expert technique to improve the expressiveness of prompt-tuning. Extensive experiments demonstrate our method is parameter-efficient, scales better with dataset size and number of prompts, and is highly modular.
{
    \small
    \bibliographystyle{ieeenat_fullname}
    \bibliography{main}
}


\end{document}